\documentclass[conference]{IEEEtran}
\IEEEoverridecommandlockouts

\usepackage[colorinlistoftodos]{todonotes}

\usepackage{cite}
\usepackage{amsmath,amssymb,amsfonts}
\usepackage{algorithmic}
\usepackage{graphicx}
\usepackage{textcomp}
\usepackage{xcolor}
\def\BibTeX{{\rm B\kern-.05em{\sc i\kern-.025em b}\kern-.08em
    T\kern-.1667em\lower.7ex\hbox{E}\kern-.125emX}}
\begin{document}

\title{Evaluating Few-Shot Pill Recognition Under Visual Domain Shift}

\author{\IEEEauthorblockN{Wai Ip Chu}
\IEEEauthorblockA{\textit{Department of Engineering} \\
\textit{City St George's, University of London}\\
London, United Kingdom \\
William.Chu@citystgeorges.ac.uk}
\and
\IEEEauthorblockN{Giacomo Tarroni}
\IEEEauthorblockA{\textit{Department of Computer Science} \\
\textit{City St George's, University of London}\\
London, United Kingdom \\
Giacomo.Tarroni@citystgeorges.ac.uk}
\and
\IEEEauthorblockN{Ling Li}
\IEEEauthorblockA{\textit{Department of Engineering} \\
\textit{City St George's, University of London}\\
London, United Kingdom \\
Caroline.Li@citystgeorges.ac.uk}
}

\maketitle

\begin{abstract}
Adverse drug events are a significant source of preventable harm, which has led to the development of automated pill recognition systems to enhance medication safety. Real-world deployment of these systems is hindered by visually complex conditions, including cluttered scenes, overlapping pills, reflections, and diverse acquisition environments. This study investigates few-shot pill recognition from a deployment-oriented perspective, prioritizing generalization under realistic cross-dataset domain shifts over architectural innovation. A two-stage object detection framework is employed, involving base training followed by few-shot fine-tuning. Models are adapted to novel pill classes using one, five, or ten labeled examples per class and are evaluated on a separate deployment dataset featuring multi-object, cluttered scenes. The evaluation focuses on classification-centric and error-based metrics to address heterogeneous annotation strategies. Findings indicate that semantic pill recognition adapts rapidly with few-shot supervision, with classification performance reaching saturation even with a single labeled example. However, stress testing under overlapping and occluded conditions demonstrates a marked decline in localization and recall, despite robust semantic classification. Models trained on visually realistic, multi-pill data consistently exhibit greater robustness in low-shot scenarios, underscoring the importance of training data realism and the diagnostic utility of few-shot fine-tuning for deployment readiness.
\end{abstract}


\section{Introduction}
Adverse drug events (ADEs) remain a significant source of preventable patient harm, particularly when patients manage complex medication regimens involving multiple concurrent prescriptions \cite{b1}. Such scenarios create challenging visual environments for computational systems, as medications must be identified under conditions that differ substantially from controlled acquisition setups. Pills are often stored together in dosette boxes or pill organizers, resulting in cluttered scenes with overlapping objects and reflective surfaces \cite{b2}. These conditions introduce occlusions, ambiguous boundaries, and visual artefacts, which present substantial challenges for conventional computer vision systems developed under idealized assumptions.

Pill recognition is a fine-grained visual classification and detection task characterized by high intra-class similarity and low inter-class variability. Many pills share similar shapes, colors, and surface textures, making their distinction challenging even under controlled imaging conditions. In real-world scenarios, these challenges are exacerbated by partial visibility, specular reflections, and background variation \cite{b3}. Although modern deep learning models can achieve high accuracy with large, well-curated datasets, assembling such datasets in healthcare contexts is often costly and logistically complex \cite{b4}. These constraints motivate the adoption of few-shot learning, which adapts models using only a small number of labeled examples and has become increasingly relevant for medical and healthcare-related vision tasks.

Few-shot learning has been investigated for pill recognition through both metric-based and meta-learning paradigms. Metric-based methods construct embedding spaces that enable recognition of novel classes using similarity measures, whereas meta-learning approaches optimize models for rapid adaptation across diverse tasks. Previous research, including studies utilizing datasets such as CURE, has demonstrated promising results with these techniques under controlled conditions \cite{b5}. However, these evaluations are typically performed with training and test data from closely related distributions, often involving isolated pills, limited background variation, and minimal object interaction\cite{b17}. Consequently, reported performance primarily reflects within-domain generalization and may overstate robustness in practical deployment scenarios\cite{b18}. 

More broadly, domain generalization and cross-dataset evaluation are recognized as significant challenges in computer vision, especially for safety-critical applications\cite{b20}. Although several studies have investigated dataset bias and performance degradation under distribution shift \cite{b6}, systematic cross-dataset evaluation is uncommon in few-shot object detection and remains largely unexplored for pill recognition\cite{b19}. Most existing few-shot pill recognition studies do not evaluate how models trained on visually homogeneous data adapt to environments with clutter, occlusion, and multiple object interactions, even though these conditions are central to real-world medication handling\cite{b18}. 

This study examines few-shot pill recognition from a deployment-oriented perspective, emphasizing generalization under realistic domain shift rather than proposing new architectures or benchmarks. The goal is not to establish state-of-the-art few-shot detection performance, but to interrogate robustness, failure modes, and supervision requirements under conditions representative of real-world deployment. Our investigation reveals the following key contributions:

\begin{itemize}
\item We demonstrate that few-shot adaptation enables reliable semantic pill recognition under severe cross-dataset domain shift. Foreground classification accuracy saturates even in the 1-shot regime, despite substantial differences in scene structure, annotation style, and visual complexity. This shows that modern detectors learn transferable semantic representations for pills that generalize beyond the distributions assumed in prior few-shot benchmarks.
\item We provide evidence that training data realism, rather than dataset size or shot count alone, is a dominant factor in few-shot generalization under clutter and occlusion. Models base-trained on visually realistic, multi-pill data consistently exhibit higher robustness and lower failure rates in overlapping scenes, revealing a strong interaction between base-domain visual statistics and low-shot adaptation behavior.
\item We identify a systematic failure mode in few-shot pill recognition, whereby semantic classification remains robust while localization and recall degrade sharply under heavy overlap. This decoupling, revealed through classification-centric and error-based metrics, highlights limitations that are obscured by conventional average-precision-focused evaluations and clean benchmark settings.
\item We establish few-shot fine-tuning as a diagnostic tool for deployment readiness, showing that varying supervision levels expose stability-robustness trade-offs and diminishing returns beyond intermediate shot counts. This reframes few-shot learning not only as a data-efficient adaptation strategy but also as a method for probing domain sensitivity and supervision requirements in real-world pill recognition systems.
\end{itemize}

Together, these contributions provide critical insight into the role of data realism and evaluation protocol in the practical design and assessment of pill recognition systems for healthcare deployment.

\section{Methods}
Automated pill recognition systems are increasingly proposed for integration into healthcare workflows such as medication dispensing, adherence monitoring, and safety verification\cite{b7}. In practical settings, however, annotated training data are often scarce, and test conditions may differ substantially from those observed during model development. These discrepancies include the presence of multiple pills per image, pill overlap, visually cluttered backgrounds, reflections from pill organizers, and non-uniform lighting. Such conditions challenge conventional object detection pipelines trained under controlled imaging assumptions\cite{b8}.

This study explores few-shot learning (FSL) as a deployment-oriented adaptation and evaluation tool for pill recognition under realistic imaging conditions. Instead of comparing datasets based on absolute performance, the analysis examines how a pre-trained object detector adapts to novel pill classes with only a small number of labeled examples and evaluates this adaptation under the visual complexity characteristic of real-world medication handling. 

\subsection{Datasets}
Two datasets, CURE and MEDISEG, are employed for base training. These datasets differ substantially in visual complexity and annotation granularity. The study intentionally leverages these differences to investigate the impact of base-domain realism on few-shot adaptation under domain shift.

CURE is a large-scale pill recognition dataset comprising 8,973 images across 196 pill classes. Each image depicts a single pill instance photographed under controlled conditions, with minimal background clutter and no object overlap\cite{b5}. Annotations are provided as full-image bounding boxes, ensuring that each image corresponds to a single pill instance. Consequently, CURE emphasizes fine-grained appearance cues in visually homogeneous settings but does not expose models to multi-object interactions or occlusion during base training.

MEDISEG comprises 8,262 images spanning 32 pill classes and is designed to reflect realistic medication-handling scenarios. Images often contain multiple pill instances per frame, with pills placed in dosette boxes or on varied backgrounds. Each pill instance is annotated with an individual bounding box, enabling explicit supervision in the presence of clutter, overlap, reflections, and partial occlusion\cite{b12}. In comparison to CURE, MEDISEG offers fewer classes but substantially greater instance-level and scene-level variability.

\begin{table}[htbp]
\caption{Characteristics of the base-training datasets used in this study. CURE provides large-scale, single-instance supervision under controlled conditions, while MEDISEG emphasizes realistic multi-object scenes with instance-level annotations.}
\begin{center}
\begin{tabular}{|l|l|l|l|l|}
\hline
Dataset & Images & Classes & Instances & Annotation \\
 &  &  & per Image & Type \\ \hline
CURE & 8,973 & 196 & 1 & Full-image \\
 & & & & bounding \\ 
 & & & & box \\ \hline
MEDISEG & 8,262 & 32 & Multiple & Instance-level \\
 & & & & bounding \\
 & & & & boxes \\ \hline
\end{tabular}
\label{tab:datasets}
\end{center}
\end{table}

\begin{figure}[htbp]
\centering
\includegraphics[width=1\linewidth]{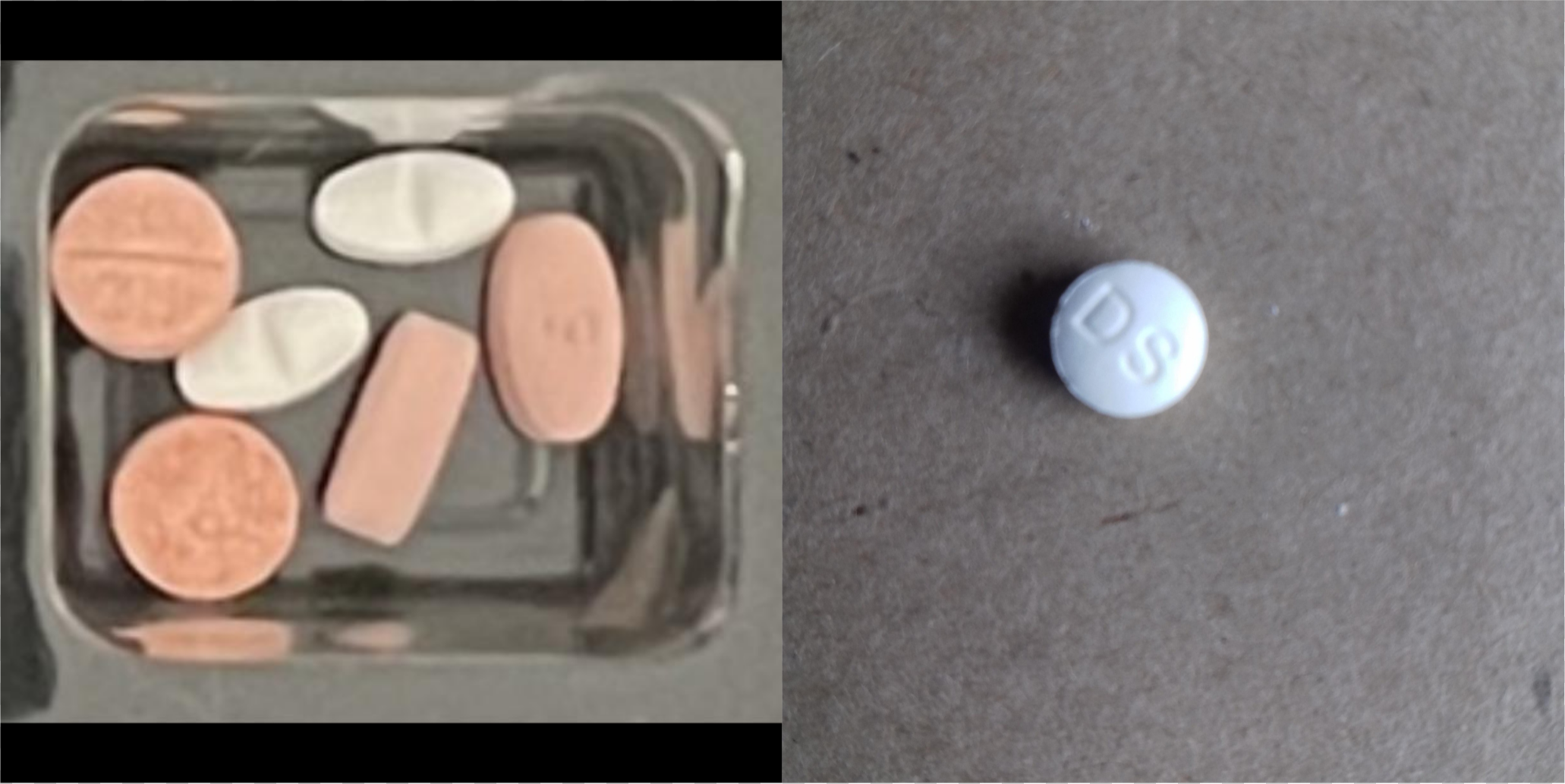}
\caption{Example images from the MEDISEG and CURE datasets. The MEDISEG image (left) contains multiple pill instances within a container, while the CURE image (right) depicts a single pill captured under controlled conditions.}
\label{fig:datasets_example}
\end{figure}

Figure~\ref{fig:datasets_example} provides a visual comparison of the MEDISEG and CURE datasets used for base training, illustrating their differing levels of visual complexity. Both datasets are used exclusively for base training and are not mixed within a single training run. Novel classes for few-shot adaptation are disjoint from base classes, and images from these datasets are excluded from the few-shot support and evaluation sets. This separation ensures that observed differences in few-shot behavior are attributable to base-domain training characteristics rather than data leakage.

\subsection{Few-Shot Learning for Object Detection}
Few-shot learning aims to adapt a model to novel categories using only a few annotated samples per class. In contrast to conventional supervised learning, which assumes abundant labeled data, few-shot learning explicitly targets low-data regimes that are common in medical and healthcare applications. In object detection, few-shot learning poses additional challenges due to the need for both localization and classification with limited supervision\cite{b9}. 

We adopt the standard N-way K-shot formulation for few-shot object detection\cite{b10}. Given:
\begin{itemize}
\item A support set $\mathcal{S} = \{(x_{i},y_{i})\}_{i=1}^{K \times N}$, containing $K$ annotated instances for each of $N$ novel classes;
\item A query set $\mathcal{Q} = \{x_{j}\}$, containing previously unseen images from the same novel classes;
\end{itemize}
The objective is to adapt a pre-trained detector using only $\mathcal{S}$ and to evaluate its performance on a disjoint query set $\mathcal{Q}$ consisting of previously unseen images containing the same novel classes.MEDISEG

\begin{figure}[htbp]
\centering
\includegraphics[width=1\linewidth]{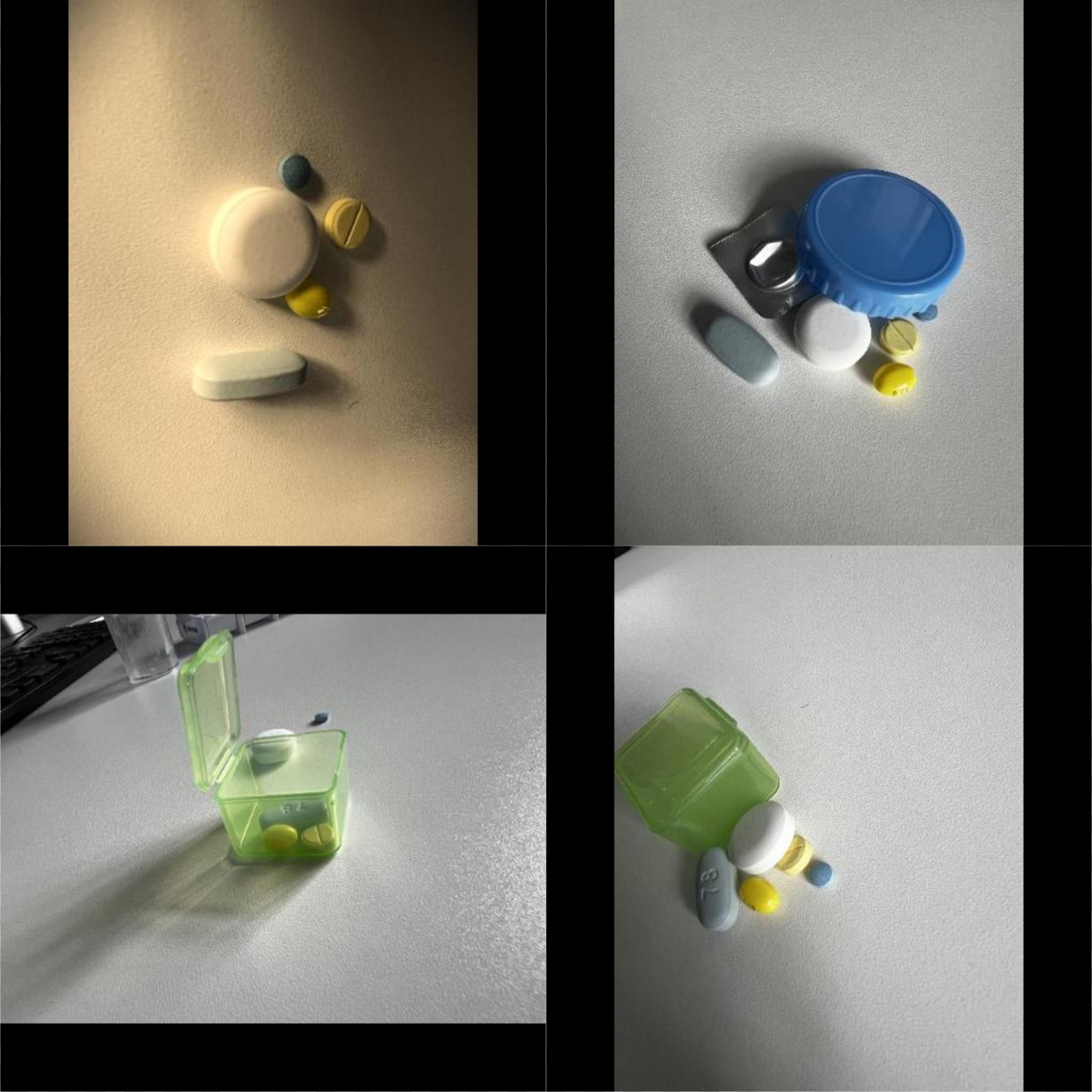}
\caption{Representative examples from the novel deployment dataset used for few-shot adaptation and evaluation. Images depict multiple pill instances with varying degrees of overlap, occlusion, and interference from pill containers and surrounding objects, reflecting realistic medication handling conditions.}
\label{fig:novel_set}
\end{figure}

In this study, few-shot adaptation is performed under a fixed 5-way setting ($N=5$), with $K \in \{1, 5, 10\}$ shots per class. For each configuration, the support set consists of $5K$ annotated instances sampled from a dedicated novel deployment dataset. Support images are excluded from all evaluation sets to prevent data leakage. Figure~\ref{fig:novel_set} illustrates representative images from this dataset, highlighting the degree of pill overlap, occlusion, and visual clutter encountered during adaptation and evaluation.

The corresponding query set comprises 516 images containing previously unseen pill arrangements, with each image including one or more pill instances. These images are not used during base training or few-shot fine-tuning and are shared across all shot settings to enable direct comparison. Query images reflect realistic deployment conditions, including cluttered scenes, overlapping pills, and partial occlusion.

All images in the novel dataset are manually annotated using the COCO Annotator tool\cite{b21}. Each pill instance is labeled with both a bounding box and an instance segmentation mask, enabling precise delineation of pill boundaries even in cases of overlap or partial visibility. These annotations support both quantitative evaluation and qualitative analysis of failure modes under realistic visual complexity.

\subsection{Few-Shot Object Detection Framework}

All experiments employ a two-stage few-shot object detection framework based on Faster R-CNN, implemented using the FsDet library \cite{b11}. Training follows the standard FsDet protocol, consisting of a base training phase on base classes and a constrained few-shot fine-tuning phase on novel classes.

\subsubsection{Base Training Phase}
During base training, the detector is trained on a set of base classes using either the CURE or MEDISEG dataset. This phase learns generic visual representations for pill appearance, region proposal mechanisms for object localization, and feature embeddings suitable for later adaptation. Let $\theta$ denote the model parameters after completion of base training. Base training settings are held fixed across all experiments to isolate the effect of few-shot adaptation.

\subsubsection{Few-Shot Fine-Tuning Phase}
Few-shot adaptation is performed by fine-tuning the base-trained detector using the support set $\mathcal{S}$ of novel classes. Fine-tuning is intentionally constrained to reduce overfitting under extreme data scarcity and to reflect realistic deployment scenarios.

Optimization is performed using stochastic gradient descent with momentum 0.9 and weight decay $1 \times 10^{-4}$. A fixed learning rate of $1 \times 10^{-3}$ is used for all few-shot configurations. Fine-tuning is carried out for a fixed budget of 2{,}000 iterations across all shot settings (1-shot, 5-shot, and 10-shot), with no early stopping, ensuring that observed performance differences arise from supervision level rather than training duration. Mini-batch sizes are kept small, consistent with the limited size of the support set.

During few-shot fine-tuning, the backbone network (ResNet with feature pyramid network) is frozen to preserve generic visual representations learned during base training. The region proposal network is partially trainable under the constrained learning rate, while the ROI heads responsible for classification and bounding-box regression are fully fine-tuned. The classification layer is re-initialized for the novel classes and trained exclusively using the support set.

Only support images are used during few-shot fine-tuning; base-training data are not revisited, and no additional data augmentation beyond standard Detectron2 transformations is applied. This strictly enforces the few-shot setting.

\subsubsection{Evaluation}
After fine-tuning, the adapted model parameters $\theta'$ are evaluated on a held-out query set containing unseen images of the novel classes, as well as on a separate overlap-focused stress-test set. Evaluation emphasizes classification-centric and error-related metrics, including foreground classification accuracy, foreground false negative rate, and loss-based indicators (classification, RPN, and total loss). Bounding-box average precision is reported where applicable but is not treated as the primary performance indicator due to heterogeneous annotation granularity across training sources.

\subsection{Data Partitioning and Cross-Domain Evaluation Protocol}
Cross-domain few-shot evaluation is designed to explicitly separate base-domain training from target-domain adaptation and testing. Base training is performed using either the CURE or MEDISEG dataset, while few-shot fine-tuning and evaluation are conducted exclusively on a separate novel deployment dataset. No images or classes are shared across these stages.

For each experimental configuration, the detector is first trained on base classes drawn from a single source dataset. Few-shot adaptation is then performed using a small support set of novel classes sampled from the deployment dataset, followed by evaluation on a disjoint query set drawn from the same target domain. This setup induces a controlled domain shift, in which both visual characteristics and annotation properties differ between the base and target domains.

Importantly, base training data are never revisited during few-shot fine-tuning, and support images are excluded from all evaluation sets. This strict separation ensures that observed performance reflects few-shot generalization across datasets rather than memorization or data leakage.

The objective of this protocol is not to compare datasets in isolation, but to analyze how differences in base-domain visual realism influence few-shot adaptation behavior when deployed in a more complex target environment.

\subsection{Evaluation Protocol and Metrics}

Evaluation is performed exclusively on query images that are disjoint from both base training and few-shot fine-tuning. Query images contain multiple pill instances and exhibit visual conditions characteristic of real-world medication handling, including clutter, overlap, and partial occlusion.

Standard object detection metrics such as average precision (AP) depend critically on consistent bounding-box definitions and intersection-over-union (IoU) thresholds across training and evaluation data. In this study, base-training datasets differ substantially in annotation granularity: CURE provides coarse, full-image bounding boxes for single-pill images, whereas MEDISEG and the novel deployment dataset provide instance-level annotations for multi-object scenes. These differences lead to systematic inconsistencies in localization supervision and proposal matching, making AP values across experimental configurations not directly comparable.

As a result, AP is reported only where meaningful but is not used as the primary indicator of performance. Instead, the analysis emphasizes classification-centric and error-related signals derived from the region-of-interest (ROI) head of the detector, which isolate semantic recognition behavior from localization artefacts. This choice enables fair comparison across models trained on heterogeneous annotation sources and focuses evaluation on recognition reliability under domain shift.

The following metrics are extracted during evaluation:
\begin{itemize}
\item Foreground classification accuracy\\
Measures the proportion of correctly classified foreground proposals:
\[
\mathrm{FG\text{-}Acc} =
\frac{\text{Correct foreground classifications}}
     {\text{Total foreground proposals}}
\]
\item False negative rate\\
Quatifies missed detections:
\[
\mathrm{FN} =
\frac{\text{Missed ground-truth objects}}
     {\text{Total ground-truth objects}}
\]
\item Region Proposal Network (RPN)\cite{b15} classification loss\\
Reflects the model’s ability to distinguish object-like regions from background. 
\item Total detection loss\\
Aggregates classification, localization, and proposal losses, providing a holistic measure of model stability during inference. 
\end{itemize}

Together, these metrics enable analysis of few-shot generalization behavior that is robust to annotation heterogeneity. Foreground classification accuracy reflects semantic recognition capability once an object is localized, while false negative rate and RPN loss expose recall and proposal failures that dominate under severe visual complexity. This decomposition is particularly informative in few-shot settings, where high semantic accuracy may coexist with localization instability and missed detections—failure modes that are obscured by aggregate AP scores alone.

\subsection{Overlap-Focused Stress Testing}
To explicitly probe robustness under severe visual complexity, we conduct an additional stress test using an overlap-only evaluation set composed exclusively of cluttered pill images exhibiting heavy instance overlap. These scenes include partial occlusion, close spatial proximity between different pill classes, and visual artefacts such as reflections from pill containers, representing challenging conditions commonly encountered in real-world medication handling.

The overlap-only evaluation follows the same few-shot construction protocol as the standard novel dataset. Few-shot adaptation is performed under a fixed 5-way setting with 1-shot, 5-shot, and 10-shot support configurations. Support sets are constructed using the same sampling and annotation procedure described for the primary novel dataset and are disjoint from all evaluation images.

The corresponding overlap-only query set consists of 133 images, each containing one or more pill instances with significant spatial overlap. Images are selected from the deployment dataset solely based on the presence of overlapping pill instances, independent of any model predictions, and the same selection protocol is applied across all base-training configurations. All images are manually verified to ensure substantial occlusion or boundary ambiguity and are annotated with instance-level bounding boxes and segmentation masks, enabling accurate assessment under extreme visual conditions.

Importantly, the overlap-only query set is excluded from base training, few-shot fine-tuning, and standard query evaluation. This ensures that evaluation is conducted under a controlled distribution shift that preserves the label space while altering scene structure and visual complexity, consistent with prior work on dataset generalization \cite{b16}. Quantitative evaluation is complemented by qualitative visualization of detection outputs to facilitate analysis of failure modes and generalization behavior across different few-shot configurations.

\subsection{Reproducibility}
All experiments are conducted using identical backbone architectures and optimization settings across few-shot configurations. Multiple runs are performed using different random seeds, and metrics are reported as mean and standard deviation across runs. Support and query set splits are explicitly defined and held fixed across shot settings to ensure comparability.

The methodology is implemented using open-source frameworks (FsDet and Detectron2), with dataset registration and evaluation scripts provided to facilitate independent replication. 

\section{Results and Discussion}
\subsection{Few-Shot Adaptation Under Domain Shift}
Few-shot fine-tuning is evaluated as a deployment-oriented adaptation mechanism under domain shift, where target images differ substantially from the base training distribution in terms of scene complexity, background clutter, pill overlap, and imaging conditions. Unlike canonical few-shot benchmarks, novel classes and evaluation data are drawn from a distinct target dataset, reflecting realistic deployment scenarios in which a pre-trained detector must adapt to a new environment using only a small number of labeled examples.

Due to heterogeneous annotation strategies across training sources, bounding-box-based metrics such as average precision are not directly comparable and are therefore excluded from primary analysis. Instead, we focus on classification-centric signals derived from the detector’s ROI head, including foreground classification accuracy, foreground false negative rate, and loss-based indicators. These metrics isolate semantic recognition and proposal reliability while remaining robust to annotation inconsistencies. 

\begin{figure}[htbp]
\centering
\includegraphics[width=1\linewidth]{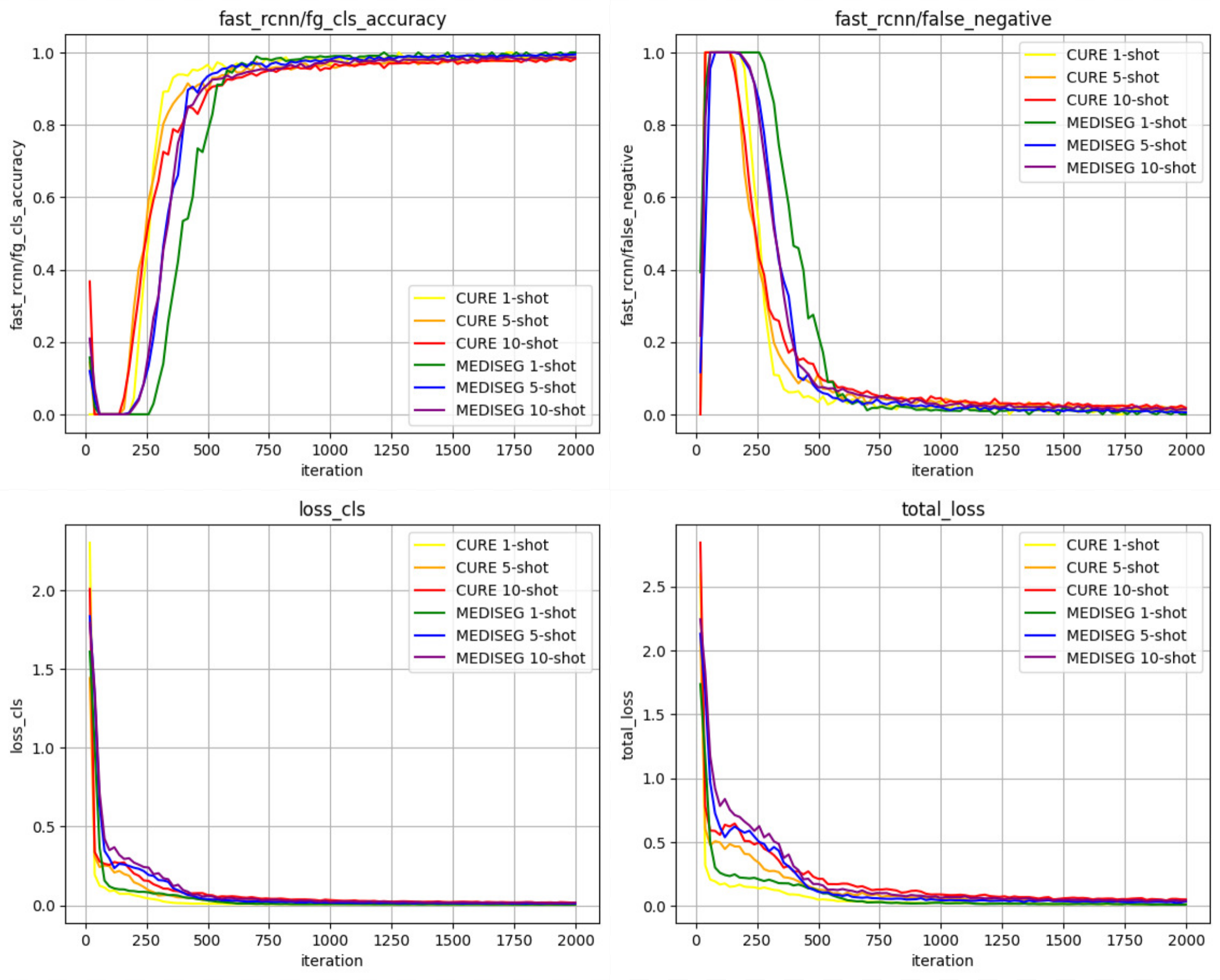}
\caption{Evolution of classification-centric and loss-based metrics during few-shot fine-tuning for CURE- and MEDISEG-trained models under 1-shot, 5-shot, and 10-shot configurations. Shown are foreground classification accuracy, foreground false negative rate, classification loss, and total detection loss over 2,000 fine-tuning iterations.}
\label{fig:metrics}
\end{figure}

Figure~\ref{fig:metrics} illustrates the evolution of classification-centric and loss-based metrics during few-shot fine-tuning for all 1-shot, 5-shot, and 10-shot configurations. The trajectories show rapid convergence of foreground classification accuracy and corresponding reductions in false negative rate across both CURE- and MEDISEG-trained models, while classification and total losses stabilize at higher values as supervision increases, reflecting a more challenging optimization landscape rather than degraded recognition.

Table~\ref{tab:metrics} summarizes these trends quantitatively by reporting the mean and standard deviation of each metric over the final 500 training iterations. Across all configurations, few-shot fine-tuning achieves high foreground classification accuracy despite substantial domain shift. In the 1-shot regime, models base-trained on CURE achieve a mean foreground classification accuracy of 0.989, while MEDISEG-trained models reach 0.994. Although the absolute difference is small, the corresponding false negative rate is reduced from 0.011 to 0.006, representing a 45\% relative reduction in missed detections. This indicates that exposure to visually realistic, multi-instance training data improves recognition reliability even under extreme supervision scarcity.

\begin{table}[htbp]
\caption{Metrics reported as mean $\pm$ standard deviation over the final 500 training iterations. Bounding-box metrics are omitted due to annotation heterogeneity across training sources; results focus on classification-centric signals from the ROI head.}
\begin{center}
\begin{tabular}{|l|c|c|c|c|}
\hline
& \textbf{fg\_cls\_accuracy} & \textbf{false\_negative} & \textbf{loss\_cls} & \textbf{total\_loss} \\ \hline
CURE & $0.989$ & $0.011$ & $0.005$ & $0.015$ \\
1-shot & $\pm 0.004$ & $\pm 0.004$ & $\pm 0.001$ & $\pm 0.003$ \\ \hline
CURE & $0.980$ & $0.020$ & $0.014$ & $0.039$ \\
5-shot & $\pm 0.004$ & $\pm 0.004$ & $\pm 0.001$ & $\pm 0.003$ \\ \hline
CURE & $0.977$ & $0.023$ & $0.019$ & $0.055$ \\
10-shot & $\pm 0.004$ & $\pm 0.004$ & $\pm 0.002$ & $\pm 0.005$ \\ \hline
MEDISEG & $0.994$ & $0.006$ & $0.005$ & $0.014$ \\
1-shot & $\pm 0.005$ & $\pm 0.005$ & $\pm 0.001$ & $\pm 0.002$ \\ \hline
MEDISEG & $0.991$ & $0.009$ & $0.011$ & $0.032$ \\
5-shot & $\pm 0.002$ & $\pm 0.002$ & $\pm 0.001$ & $\pm 0.003$ \\ \hline
MEDISEG & $0.983$ & $0.017$ & $0.015$ & $0.044$ \\
10-shot & $\pm 0.003$ & $\pm 0.003$ & $\pm 0.002$ & $\pm 0.003$ \\ \hline
\end{tabular}
\label{tab:metrics}
\end{center}
\end{table}

Although classification and total losses increase with higher shot counts, these increases do not reflect degraded recognition performance. For CURE-trained models, total loss increases from 0.015 in the 1-shot regime to 0.055 at 10-shot (+267\%), while foreground classification accuracy decreases by only 1.2 percentage points (0.989 to 0.977). Similar trends are observed for MEDISEG-trained models. This divergence indicates that additional supervision introduces a more challenging optimization landscape without substantially affecting semantic recognition, and that the detector converges reliably across all shot settings.

Overall, these results demonstrate that few-shot fine-tuning enables rapid adaptation to novel pill classes under severe domain shift, with semantic recognition saturating early and remaining stable across base training sources.

\subsection{Effect of Supervision Level}
The effect of supervision level is examined by comparing 1-shot, 5-shot, and 10-shot fine-tuning regimes. Rather than seeking monotonic improvements in mean accuracy, this analysis focuses on how additional labeled support examples influence robustness, variance, and optimization stability.

As summarized in Table~\ref{tab:supervision-level}, foreground classification accuracy is already high in the 1-shot regime for both CURE- and MEDISEG-trained models, indicating early saturation of semantic recognition. Increasing supervision does not consistently improve mean accuracy and, in some cases, leads to slight reductions. For example, MEDISEG-trained models exhibit a decrease from 0.994 at 1-shot to 0.983 at 10-shot, a change of 1.1 percentage points, consistent with stochastic effects and class similarity in few-shot detection.

\begin{table}[htbp]
\caption{Foreground classification accuracy and false negative rate are reported as mean ± standard deviation over the final 500 training iterations for 1-, 5-, and 10-shot fine-tuning. Metrics are grouped by shot count. }
\begin{center}
\begin{tabular}{|l|c|c|c|c|}
\hline
 & \textbf{fg\_cls\_} & \textbf{fg\_cls\_} & \textbf{false\_} & \textbf{false\_} \\
 & \textbf{accuracy} & \textbf{accuracy} & \textbf{negative} & \textbf{negative} \\
 & \textbf{(CURE)} & \textbf{(MEDISEG)} & \textbf{(CURE)} & \textbf{(MEDISEG)} \\ \hline
1-shot & $0.989$ & $0.994$ & $0.011$ & $0.006$ \\
& $\pm 0.004$ & $\pm 0.005$ & $\pm 0.004$ & $\pm 0.005$ \\ \hline
5-shot & $0.980$ & $0.991$ & $0.020$ & $0.009$ \\
& $\pm 0.004$ & $\pm 0.002$ & $\pm 0.004$ & $\pm 0.002$ \\ \hline
10-shot & $0.977$ & $0.983$ & $0.023$ & $0.017$ \\
& $\pm 0.004$ & $\pm 0.003$ & $\pm 0.004$ & $\pm 0.003$ \\ \hline
\end{tabular}
\label{tab:supervision-level}
\end{center}
\end{table}

In contrast, additional supervision yields clear gains in training stability. For MEDISEG-trained models, increasing supervision from 1-shot to 5-shot reduces the standard deviation of foreground classification accuracy by 60\% (±0.005 to ±0.002), while variance in loss-based metrics is similarly reduced. Although foreground false negative rates do not decrease monotonically across all settings, their variability decreases with additional shots, indicating improved robustness.

From a deployment perspective, these findings suggest diminishing returns beyond minimal supervision. Intermediate supervision levels, such as 5-shot, capture the majority of stability gains while avoiding the limited additional benefit observed at 10-shot, offering a practical balance between annotation effort and reliability.

\subsection{Evaluation of Overlapping and Cluttered Pill Images}
The overlap-only stress test constitutes a critical component of this study, as it isolates the most challenging visual conditions encountered in real-world medication handling. By restricting evaluation to cluttered scenes with overlapping pill instances, this analysis explicitly probes robustness under heavy occlusion, ambiguous object boundaries, and region proposal ambiguity—failure modes that are largely suppressed in standard benchmarks.

Performance degrades substantially in low-shot regimes for all models; however, pronounced differences emerge depending on the visual realism of the base training data. Table~\ref{tab:overlap} summarizes results across 1-shot, 5-shot, and 10-shot configurations. For models base-trained on CURE, foreground classification accuracy collapses from 0.989 under standard evaluation to 0.131 in the 1-shot overlap setting, representing an 87\% relative decrease. In contrast, MEDISEG-trained models retain a foreground classification accuracy of 0.406 under identical conditions, yielding a 210\% relative improvement over CURE-trained models (0.406 vs. 0.131).

This advantage persists consistently across all supervision levels. In the 5-shot regime, MEDISEG-trained models achieve a foreground classification accuracy of 0.625 compared to 0.372 for CURE-trained models, corresponding to a 68\% relative improvement. At 10-shot, the gap remains substantial, with MEDISEG-trained models reaching 0.740 versus 0.558 for CURE-trained models (33\% relative improvement). Notably, the relative advantage of MEDISEG-trained models is largest in the most data-scarce settings, indicating that exposure to realistic multi-object interactions during base training is particularly beneficial when supervision is limited.

In addition to improved recognition accuracy, MEDISEG-trained models exhibit systematically lower false negative rates and reduced region proposal and classification losses across all overlap configurations. For example, in the 1-shot overlap setting, MEDISEG reduces the false negative rate by 37\% relative to CURE (0.513 vs. 0.816), indicating improved instance recall under severe occlusion. These reductions are accompanied by lower RPN and total losses, suggesting more reliable region discovery in cluttered scenes.

Taken together, the overlap stress test provides strong empirical evidence that base-domain visual realism is a dominant factor in few-shot generalization under severe visual ambiguity. While increasing supervision improves performance for both training regimes, models trained on realistic multi-object data consistently maintain a substantial robustness advantage, particularly in low-shot settings. This finding underscores the importance of dataset design and visual complexity during base training when deploying few-shot pill recognition systems in real-world environments.

\begin{table}[htbp]
\caption{Few-shot evaluation on an overlap-only test set. Metrics emphasise foreground recognition and error behaviour under realistic visual complexity. In the 1-shot regime, models base-trained on MEDISEG achieve up to a 210\% relative improvement in foreground classification accuracy compared to CURE.}
\begin{center}
\begin{tabular}{|l|c|c|c|c|c|}
\hline
 & \textbf{fg\_cls\_} & \textbf{false\_} & \textbf{loss\_} & \textbf{loss\_rpn\_} & \textbf{total\_} \\
 & \textbf{accuracy} & \textbf{negative} & \textbf{cls} & \textbf{cls} & \textbf{loss} \\ \hline
CURE & & & & & \\
1-shot & 0.131 & 0.816 & 0.351 & 0.863 & 1.326 \\ \hline
CURE & & & & & \\
5-shot & 0.372 & 0.465 & 0.421 & 0.224 & 0.844 \\ \hline
CURE & & & & & \\
10-shot & 0.558 & 0.342 & 0.320 & 0.133 & 0.674 \\ \hline
MEDISEG & & & & & \\
1-shot & 0.406 & 0.513 & 0.383 & 0.312 & 0.963 \\ \hline
MEDISEG & & & & & \\
5-shot & 0.625 & 0.246 & 0.279 & 0.182 & 0.680 \\ \hline
MEDISEG & & & & & \\
10-shot & 0.740 & 0.210 & 0.191 & 0.059 & 0.445 \\ \hline
\end{tabular}
\label{tab:overlap}
\end{center}
\end{table}

Increasing supervision yields substantial gains under overlap, particularly between 1-shot and 5-shot. For MEDISEG-trained models, foreground classification accuracy increases from 0.406 to 0.625 (+54\% relative improvement), while false negative rate decreases from 0.513 to 0.246 (52\% reduction). Gains from 5-shot to 10-shot are smaller, with accuracy increasing by 18\%, indicating diminishing returns beyond intermediate supervision levels.

Across all shot settings, MEDISEG-trained models consistently outperform CURE-trained counterparts, achieving between 33\% and 210\% higher foreground classification accuracy under overlapping conditions. This performance gap widens in lower-shot regimes, demonstrating that exposure to realistic multi-object scenes during base training is critical for robustness under severe visual ambiguity.

\subsection{Qualitative Analysis}
\begin{figure}[htbp]
\centering
\includegraphics[width=1\linewidth]{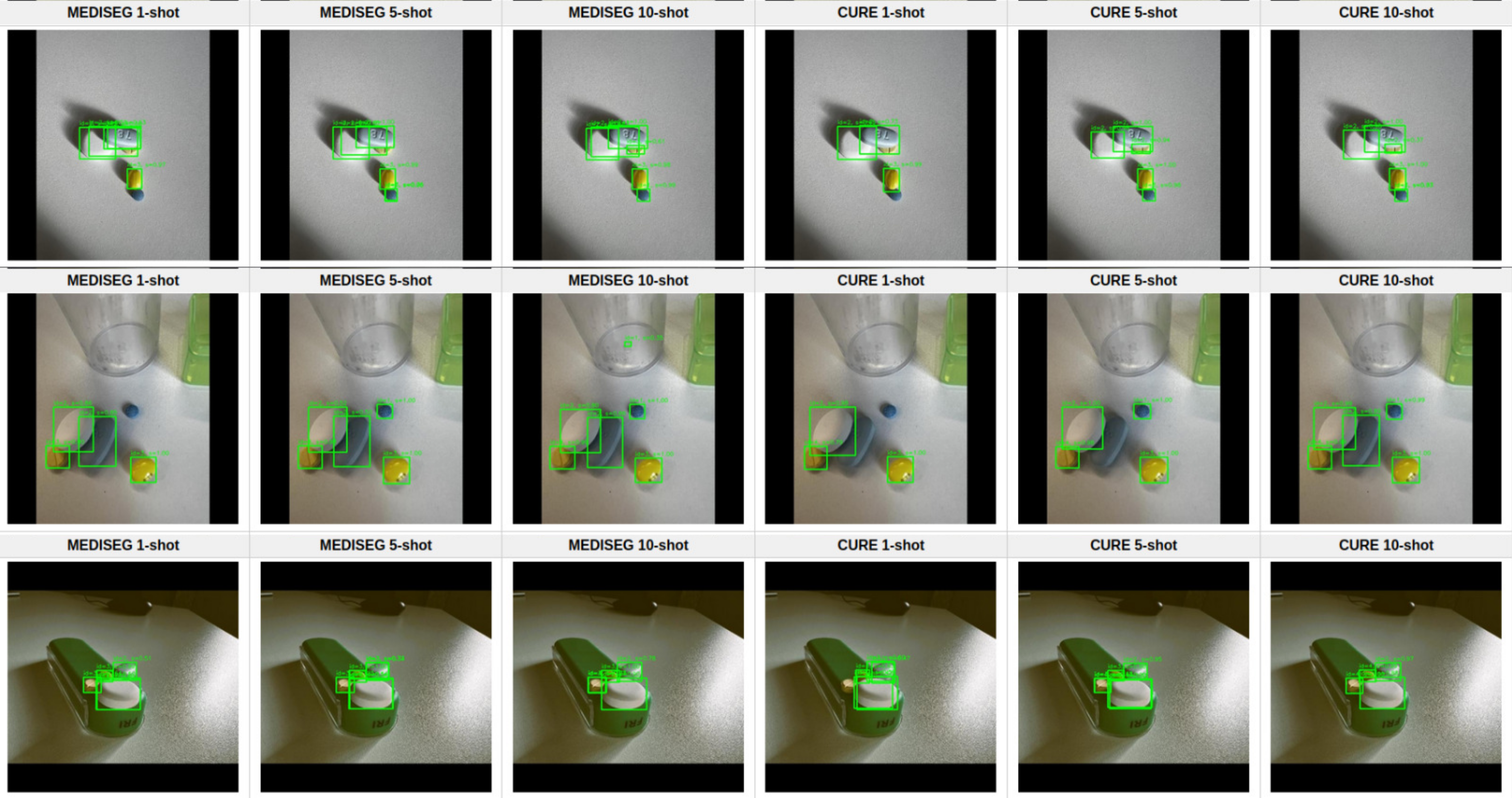}
\caption{Qualitative comparison of few-shot detection results for models base-trained on MEDISEG and CURE under 1-shot, 5-shot, and 10-shot fine-tuning. Each row shows a different test image, and each column corresponds to a specific base-training dataset and supervision level. Predicted bounding boxes illustrate detection and classification behavior under clutter, overlap, and occlusion.}
\label{fig:vis_fsl}
\end{figure}

Qualitative inspection of model predictions provides further insight into few-shot behavior under visual clutter. Figure~\ref{fig:vis_fsl} show that low-shot configurations frequently miss instances in heavily occluded regions, particularly for models base-trained on CURE. These qualitative failures align with quantitative results, where MEDISEG-trained models achieve up to 3.1× higher foreground classification accuracy than CURE-trained models in the 1-shot overlap setting.

Despite localization failures, semantic misclassification is comparatively rare once objects are detected. As supervision increases, predictions become more spatially consistent, with improved instance separation and fewer missed detections. However, failure cases persist in regions of extreme overlap, reinforcing the observed decoupling between semantic recognition and localization performance.

\subsection{Discussion and Limitations}
Several limitations of this study should be acknowledged. First, the use of full-image bounding boxes in CURE restricts the applicability of localization-sensitive metrics and necessitates reliance on classification-centric evaluation. Second, the evaluation protocol departs from canonical few-shot benchmarks by intentionally introducing cross-dataset domain shift, precluding direct benchmark comparison. Third, the number of novel classes is limited by practical annotation constraints.

Importantly, these limitations reflect real-world deployment conditions rather than benchmark artefacts. In operational medication recognition systems, annotation heterogeneity, visual domain shift, and limited supervision are the norm. Within this context, few-shot fine-tuning serves not only as an adaptation mechanism but also as a diagnostic tool for exposing failure modes and supervision requirements. The pronounced sensitivity to occlusion and diminishing returns beyond intermediate supervision levels underscore challenges that are unlikely to be revealed by clean, homogeneous benchmarks alone.

\section{Conclusion}
This work investigated few-shot learning as a deployment-oriented adaptation and evaluation framework for pill recognition under realistic visual conditions. By framing few-shot fine-tuning as both an adaptation strategy and a diagnostic tool, the study examined how object detection models respond to severe domain shift introduced by cluttered scenes, overlapping instances, and heterogeneous annotation strategies. Rather than emphasizing absolute benchmark performance, the analysis focused on classification-centric and error-related signals that more directly reflect recognition reliability in practical settings.

Across all few-shot configurations, the results demonstrate that modern object detectors can rapidly adapt to novel pill classes with extremely limited supervision, achieving high foreground classification accuracy even in the 1-shot regime. This indicates that semantic representations learned during base training transfer effectively to new pill categories despite substantial domain shift. However, increased visual complexity, particularly heavy overlap and occlusion, exposes pronounced limitations in localization and recall, revealing failure modes that are not apparent in visually simplified benchmarks and are obscured by aggregate detection metrics.

A central finding of this study is that the visual realism of base training data strongly influences few-shot generalization behavior. Models trained on data containing realistic multi-object interactions consistently exhibit greater robustness under cluttered and overlapping conditions, with the largest advantages observed in low-shot regimes. While increasing supervision does not guarantee monotonic improvements in all metrics, moderate supervision levels improve optimization stability and reduce variance, highlighting a practical trade-off between annotation effort and robustness.

Several limitations should be acknowledged. Heterogeneous annotation strategies across datasets restrict the uniform application of localization-sensitive metrics, necessitating a focus on classification-centric evaluation. In addition, the number of novel classes and deployment environments considered is constrained by practical annotation availability. These limitations reflect real-world healthcare settings, where data heterogeneity and limited supervision are common.

Overall, this study demonstrates that few-shot learning provides a valuable framework for both adapting and diagnosing pill recognition systems under realistic deployment conditions. The findings underscore the importance of training data realism and evaluation protocols that prioritize diagnostic insight, offering guidance for the development of robust pill recognition systems for safety-critical healthcare applications.

\bibliographystyle{IEEEtran}
\bibliography{IEEEabrv,mybibfile}

\end{document}